%% file: main.tex
\documentclass{article}

\usepackage[nonatbib, final]{neurips_2019}

\usepackage[utf8]{inputenc}
\usepackage{graphicx}
\newtheorem{prop}{Proposition}
\usepackage{amsmath,amssymb}
\usepackage{color}
\usepackage{subcaption}
\usepackage{algpseudocode}
\usepackage{algorithm}
\usepackage[export]{adjustbox}

\usepackage{booktabs}       
\usepackage{amsfonts}       
\usepackage{nicefrac}       
\usepackage{microtype}      

\usepackage{color}
\newcommand{\pluseq}{\mathrel{+}=}

\graphicspath{ {./images/} }

\title{Emergent Communication with World Models}
\author{%
  Alexander I. Cowen-Rivers\thanks{Completed during an internship at Preferred Networks} \\
  Huawei R\&D London  \\ \texttt{alexander.cowen-rivers@huawei.com} \\
\And Jason Naradowsky \\ Preferred Networks \\
  \texttt{narad@preferred.jp} \\
 \date{August 2019}}
\begin{document}

\maketitle

\input{00_abstract.tex}
\input{01_intro.tex}

\input{02_world_models.tex}

\input{03_method.tex}

\input{04_losses.tex}
\input{05_experiments.tex}
\input{06_related.tex}
\input{07_conclusion.tex}

\bibliographystyle{IEEEtran}
\bibliography{refs}

\input{appendix.tex}

\clearpage

\end{document}

%% file: 00_abstract.tex
\begin{abstract}
We introduce Language World Models, a class of language-conditional generative model which interpret natural language messages by predicting latent codes of future observations.  This provides a visual grounding of the message, similar to an enhanced observation of the world, which may include objects outside of the listening agent's field-of-view. We incorporate this ``observation'' into a persistent memory state, and allow the listening agent's policy to condition on it, akin to the relationship between memory and controller in a World Model. We show this improves effective communication and task success in 2D gridworld speaker-listener navigation tasks. In addition, we develop two losses framed specifically for our model-based formulation to promote positive signalling and positive listening. Finally, because
messages are interpreted in a generative model, we can visualize the model beliefs to gain insight into how the communication channel is utilized.
\end{abstract}

%% file: 01_intro.tex
\section{Introduction}

With the advent of deep reinforcement learning, there has been a resurgence of interest in situated emergent communication (EC) research~\cite{das2017learning,lazaridou2016multi,kottur2017natural, jaques2019social,havrylov2017emergence}.  
However, there remain many open design decisions, each of which may significantly bias the nature of the constructed language, and any agent policy which makes use of it.  

In this work, we consider how incoming messages are integrated into a listening agent's policy.  A common approach to this decision is to concatenate the message and state observation together, or to pass the message to the agent policy directly~\cite{lowe2019pitfalls,foerster2016communication}. 
But what type of communication does this design decision cater to?  We contend that this approach arguably gives the speaker too great an influence in shaping the listener's policy, potentially allowing the speaker to utilize the communication channel to encode an optimal policy. Auxiliary losses that pressure the listening agent to consistently alter its short-term behaviour in response to messages (e.g., causal influence of communication, CIC loss ~\cite{lowe2019pitfalls,jaques2019social,eccles:biases}) may further bias the agent towards this behaviour. In this situation, messages become merely commands.  

In contrast, we aim to develop agents that utilize information consistently, regardless of whether that information was obtained from their observations or via communication with other agents. By constraining the agent in this way, we aim to shape the use of the communication channel towards conveying more substantive information.  
We formalize this guiding assumption as follows:

\begin{quote}
In a partially-observed setting, let $o$ be an observation beyond the listener's field-of-view. Assuming the speaker is as reliable as the listener's perception, the listener's actions upon receiving full information regarding $o$ should be identical to those if given the perceptual ability to observe $o$ themselves.
\end{quote}


Therefore, we propose to explicitly separate the process of message interpretation from the agent's decision-making.  
By forcing the agent to ground incoming messages prior to taking action, we decrease the odds of the message directly controlling the agent's policy. However, accomplishing this raises a difficult modelling problem: how does an agent learn to ground messages which refer to objects outside their field of view at the time when that message is received?

We introduce \emph{Language World Models} (LWMs), an adaptation of world models~\cite{ha2018worldmodels} to partially-observable worlds which are trained to predict future states based on messages. This, in turn, creates a visual grounding for agents to exploit: upon receiving a message, an agent can immediately ``visualize'' information beyond its field-of-view and act accordingly. These visualizations begin as a reflection of entire trajectories, but refine their focus 
and may eventually hone in on the intended message semantics if the model is exposed to sufficient variation in the environment.

Our contributions are the following: 
\begin{enumerate}
    \item 
    We introduce the LWM and apply it to a 2D gridworld EC task, in which an all-seeing speaking agent must guide a listening agent to a goal location. We show that agents trained with access to a LWM exhibit greater success in this domain than those which do not, especially in sparse-messaging, longer-trajectory scenarios.  
    \item Reflecting on the aforementioned experiments, we discuss necessary conditions for successfully training a LWM.
    \item By virtue of this, we provide a solution to the long-standing challenge of how we can interpret or evaluate the listener's understanding of incoming messages (see ~\cite{lowe2019pitfalls} for an overview).
    \item We provide two additional losses formulated around this model-based approach, one to promote efficient speaking, the other to promote effective listening and show they accelerate learning along both dimensions.  
\end{enumerate}

%% file: 02_world_models.tex
\section{Language World Models}
\begin{figure}
\centering
\begin{minipage}{.44\textwidth}
  \centering
  \includegraphics[width=.8\linewidth]{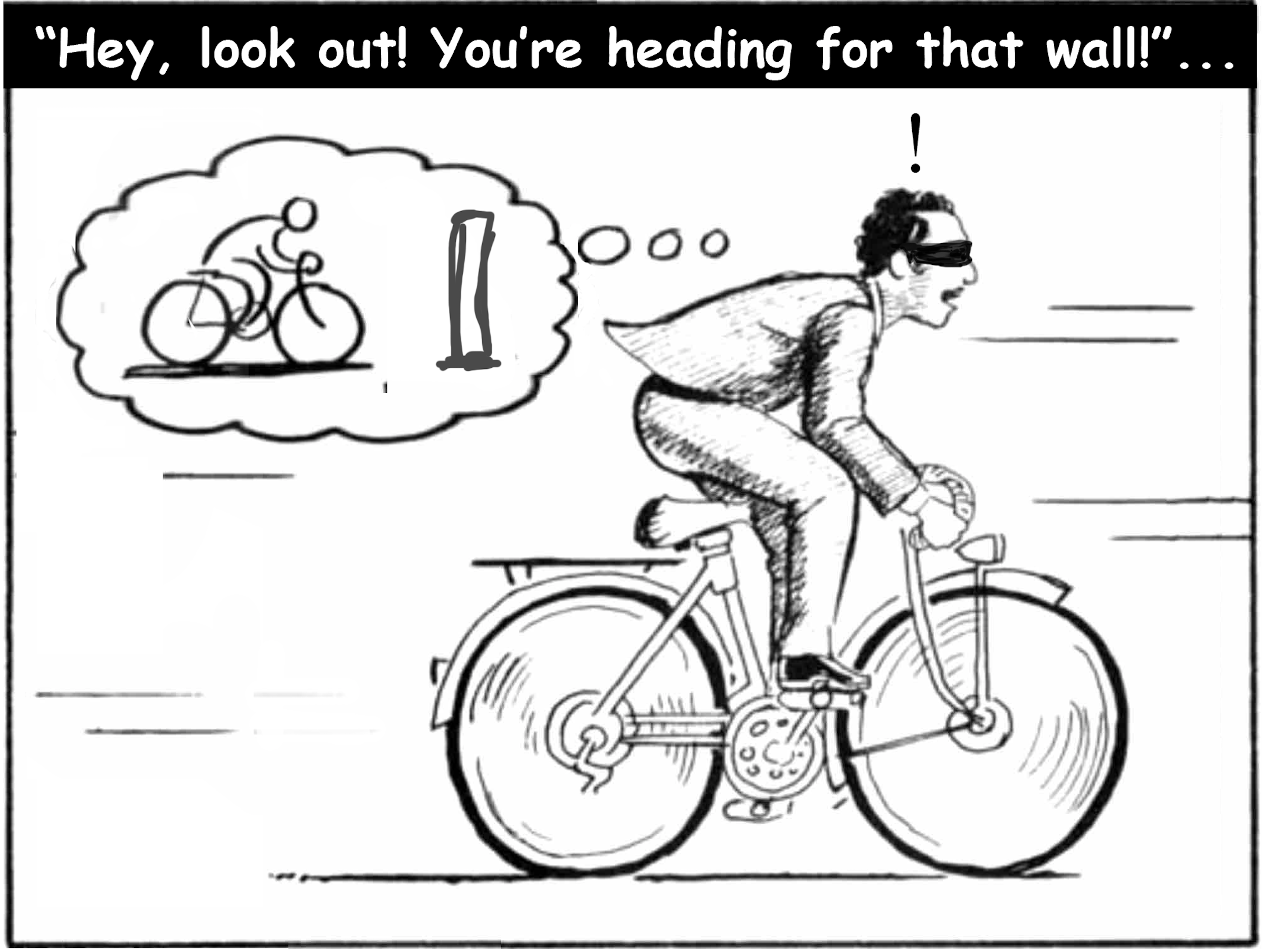}
  \captionof{figure}{A \emph{Language-Conditional} World Model, adapted from Scott McCloud’s \emph{Understanding Comics}~\cite{mccloud1998understanding} and World Models~\cite{ha2018world}.
Here the cyclist has limited observability of the world around him (blindfolded), and conceptualizes danger by interpreting language within the context of his world model.}\label{fig:lcwm-concept}
\end{minipage} \hfill 
\begin{minipage}{.44\textwidth}
  \centering
  \includegraphics[width=1.0\linewidth]{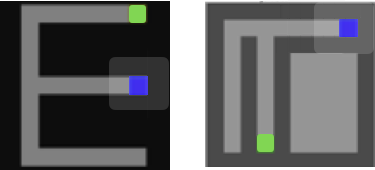}
  \captionof{figure}{The partially-observable worlds that the agents interact in. The speaker (unseen) is able to view the entire map, whereas the listener (blue) only views a pixel in each direction. At the start of each game, a flag (green) is randomly placed in one of two paths.  Both the speaker and listener receive a reward if the listener is able to find the flag by choosing the correct corridor.}\label{fig:game-simple}    \vspace{14pt}

\end{minipage}
\end{figure}

Revisiting the motivating principles of World Models~\cite{ha2018worldmodels}, we return to the following quote from Jay Wright Forrester:
\begin{quote}
“The image of the world around us, which we carry in our head, is just a model. Nobody in his head imagines all the world, government or country. He has only selected concepts, and relationships between them, and uses those to represent the real system.”
\end{quote}

\noindent To this end, a world model (WM) aims to identify such important concepts through their ability to predict future states.  
Formally, given a latent code $z_t$ for time $t$, a WM models a future state as $P(z_{t+1} | z_t, a_t)$, the outcome of action $a_t$\footnote{As described in \cite{ha2018worldmodels}, it is also possible for the WM to be recurrent, and to further condition this prediction on hidden state $h_t$.}.  Thus, a world model is useful because it presents to the agent a glimpse into a future world, beyond its current sights.

A similar problem arises in partially-observable worlds, where important concepts lie not in a possible future, but outside of the agent's field-of-view. Here we must reconsider the modelling objective, and swap short-term dependencies in time, for potentially long-term dependencies in space. 
At a given time $t$, let $O_{t} \in \mathcal{R}^N$ be the observation by an omniscient agent (speaker), $m_t$ be a message sent by that agent, and $o_{t} \in \mathcal{R}^N$ be a partial observation (of the listener) of $O_{t}$. A straightforward model for visually grounding the message is $P(O_{t} | m_t, o_{t})$\footnote{A reference to the listener's observation $o_{t}$ is important in situations where the speaker is using their knowledge of the listener's location and orientation when deciding what to communicate.}. The disadvantage of this formulation is that estimating it directly from  $O_t$,$m_t$ pairs violates important assumptions of emergent communication, making it unsuitable for our use purposes.  

We instead aim to approximate this by using the message to predict future states along the listening agent's trajectory.  As in WM's, we project an observation $o_t$ into a latent code $z_{t} \in \mathcal{R}^n$, $n\ll N$. We introduce \emph{Language World Models} as:

\begin{equation}
    P( z_{ \{ l : l \gneq t  \}} |m_t,z_t) = P(z_{t+1}, \ldots , z_T | m_t, z_t)
\end{equation}

Where ${ \{ l : l \gneq t  \}}$ is the set of future time-points greater than t. LWMs (Fig~\ref{fig:lcwm-concept}) provides useful information for agent planning by interpreting incoming messages, and passing the resulting latent code to the agent as a form of enhanced observation -- one that contains important objects both within its field of vision, as well those in more distant locations.

This formulation has connections to previous work, most notably other generative models of the world.  Table~\ref{table:definitions} provides an overview.
In comparison to WM's, the predictive power of $a_t$ is replaced by $m_t$.  Conceptually we are replacing ``\emph{What do I expect will happen if I do this?}'' with ``\emph{What do I expect to find if I hear this?}''. But as the WM conditions on $a_t$ and models purely local phenomena ($t+1$), it captures the physics of the world in ways that the LWM does not. Instead the LWM has a broader temporal window, as found in a temporal difference variational auto-encoder~\cite{gregor2018temporal}, but differs in that we do not know or specify how far in the future to predict.  


\begin{table}[t]
  \centering
  \begin{tabular}{lll}
    \toprule
    \multicolumn{3}{c}{World Modelling Components}                   \\
    \cmidrule(r){1-3}
    Model Name     & What does it model?     & Predictive PDF \\
    \midrule
    WM~\cite{ha2018worldmodels,hafner2018learning}  & Latent transitions from actions.      & $P(z_{t+1}|a_t,z_t)$ \\
    TD-VAE~\cite{gregor2018temporal}     & Long range dependencies.     & $P(z_{t_2} |z_{t_1})$ \\
    LWM (\emph{Ours})    & Latent language semantics.     & $P( z_{ \{ l : l \gneq t  \}} |m_t,z_t)$ \\
    \midrule
    Reward Model~\cite{ha2018worldmodels,hafner2018learning}      & Reward from latent transitions.   & $P(r_{t} |z_t)$ \\
    Observation Model~\cite{ha2018worldmodels,hafner2018learning}     & Observation from latent transitions.     & $P(o_{t} |z_t)$ \\
    Concept Clustering (\emph{Ours})    & Observation from messages.     & $P(o_{t} |  m_t)$ \\ 
    \bottomrule \\
  \end{tabular}
\caption{A comparison of various world models.}
\label{table:definitions}
\end{table}

\subsection{Guiding Assumptions}
Several important guiding assumptions must hold for the effective training of a LWM in the EC setting, where supervision comes only from the task loss.

\paragraph{Trustworthiness} LWMs makes strong assumptions linking the task reward to the grounding of the message: it is assumed that the information conveyed in the message is important for the completion of the task, and that therefore the agent is more likely to observe the intended target of the message along trajectories for which it receives high reward than those which it does not.  Thus, it is assumed that the speaker is trustworthy, and aims to send useful information.

\paragraph{Contrast and Consistency}
Second, while we do not know which observation(s) the speaker is trying to communicate, it is assumed that the observations which correspond to the true target of the message will, over time, be observed more often than those which do not.  While we show that the LWM is effective even when there are only two possible trajectories (one resulting in success, one in failure), a LWM undesirably captures both the important objects and closely-related (but unimportant) states. Only over many trials and permutations of the environment will the intended semantics of the message emerge.

\paragraph{Object Permanence} A final constraint in our current formulation is that objects must remain fixed across time.  This allows an observation at $o_{t+m}$, some arbitrary $m$ steps into the future, to provide an understanding about what the speaker implied at time $t$.


%% file: 03_method.tex
\section{Architecture}
Returning to the task of emergent communication, we now provide an overview of our system.

\subsection{Model of the Listening Agent}\label{sec:listagent}

The proposed modelling contributions which involve the LWM are contained within the listening agent. We describe this agent's components in terms of the WM categories defined in \cite{ha2018worldmodels}: a vision network (V), a memory network (M) and a controller (C).

\paragraph{VAE-Seq (V)}
The environment produces an observation $O_t$ at each step $t$.  In our case $O_t$ is a 2D image from a grid-world game. From this observation, the listening agent observes a partial view, $o_t$.  The VAE-Seq component compresses the observation $o_t$ into a latent code $z_t$.  We use a simple 2D convolutional network for this component.  

\paragraph{Latent Belief Network (M)}
The role of the Latent Belief network (LBF) is to compress the latent codes across the time axis, and to tie this representation to a message.  Here a message is a sequence of discrete tokens. For each sequence of observations, when a message is sent at $t = i$, we take the pair $(m_i, z_i)$,
and use this infer all future observations $z_{i+1}, . . . , z_T$.  

\paragraph{Controller (C)}
The controller takes an action $a_t$ based on the input features $z_t$, and current belief state. Following previous work \cite{lowe2019pitfalls}, we model the agent policy using a feed-forward network, which we train via
REINFORCE~\cite{williams1992simple}.

\begin{figure}[t]
  \centering
\includegraphics[width=0.7\linewidth]{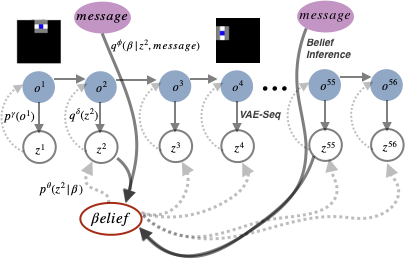}
\caption{A depiction of how the listener agent grounds incoming messages. Where $q^{\delta}(z)$ is the variational approximation to $P(z|o)$ and $q^{\phi}(\beta \mid z,m)$ is the variational approximation to $P(\beta \mid z,m)$. See Sec~\ref{sec:listagent} for a reminder of the listener agents derivation.}
\label{fig:bmodel}
\end{figure}

\paragraph{Listener Agent Summary}
$P(z|o)$ compresses the observation, then $P( z_{ \{ l : l \gneq t \} } \mid m,z)$ grounds the message with future compressed observations --- we further decompose the model $P( z_{ \{ l : l \gneq t \} } \mid m,z)$ by assuming there exists a latent variable $\beta$ (persistent memory) such that $P( z_{ \{ l : l \gneq t \} } |m,z) = \int_{\beta} P( z_{ \{ l : l \gneq t \} } \mid \beta) P(\beta \mid m,z)$.  Where we define the specific case where no message is received as $P(\beta \mid m=None,z=z_t) = \beta_{t-1}$, otherwise the model uses amortised VI~\cite{kingma2013auto} to estimate $\beta$. Lastly $P(a|z,\beta)$ selects an action conditioned on $\beta$ and compressed observation 

\subsection{Model of the Speaking Agent}

The speaker agent takes in the global observation $O_t$ and produces a message $m_t$.  In this work, the speaker is modelled using a convolutional network, linear layer, and softmax to produce a 1-hot representation of $O_t$, which becomes the message $m_t$.

%% file: 04_losses.tex
\section{Promoting Effective Communication}

Previous work~\cite{lowe2019pitfalls} defines an effective communication strategy as one that achieves \textit{positive signalling} and \textit{positive listening}. We detail these metrics and introduce a loss function that explicitly/ implicitly promotes each. 

\subsubsection{Listener}

An agent exhibits positive listening if the distribution over its actions is correlated with the messages it receives, i.e., that messages influence the agent's behaviour. A previously-proposed method for biasing agents towards positive listening is the causal influence of communication (CIC) loss\cite{lowe2019pitfalls, jaques2019social}. We describe a variant of this loss below:

\begin{algorithm}[h] 
\label{alg:cic}
\caption{One-step causal influence of communication~\cite{lowe2019pitfalls,jaques2019social}}
\begin{algorithmic}[1]
\State Agent1 policy $\pi_1$, Agent2 policy $\pi_2$ 
\State messages $\bar{m}=\{m_0, ..., m_{M-1}\}$
\State Number of test games $T$.
\State $\text{CIC} = 0$
\For{$i \in \{0, ...,  T - 1\}$} \State $\triangleright$ Generate new state $S$, observations $O$.
 \State $\triangleright$ Intervene by changing message $m_j$ 
\For{$j \in \{0, ...,  M - 1\}$} 
\State $p(m_j) \gets \pi_2(m_j|o_2)$, 
\State \hspace{10mm} $p(a | m_j) \gets \pi_1(a | o_1, m_j)$
\State $p(a, m_j) = p(a | m_j) p(m_j) $
\State $p(a) = \sum_{m \in \mathcal{A}^m} p(a, m)$
 \EndFor 
\EndFor 
\State $\text{CIC} \mathrel{+}= 1/T \cdot \sum_{a \in \mathcal{A}^e} p(a, m_j) \log \frac{p(a, m_j)}{p(a)p(m_j)}$
\end{algorithmic}
\end{algorithm}

While this loss has been shown empirically to be effective in improving communication, it is difficult to apply such intuition to natural language. Humans process vast collections of new information daily, but seldom use it to adjust their immediate behaviour. In contrast, we would like to push different messages to have separate groundings, and allow the agent to act on them \emph{accordingly}. In theory, this might entail taking a particular action many time steps later, or even ignoring the message completely.

\begin{prop}
Our formulation of the LWM yields the following optimisation problem. 

\begin{equation} \label{eq:elboA}
\begin{array}{cc}
     \max_{\theta, \rho, \phi, \delta, \gamma} & J(\rho) + \sum_{o \in \mathcal{D}} \mathbb{E}_{z \sim q^{\delta}} \left[ \log p^{\gamma}(o \mid z) \right] - \mathrm{KL}[{ q^{\delta}(z ) } \mid\mid{ p^{\gamma}(z) }] \\ 
     &+  \mathbb{E}_{\beta \sim q^{\phi}} \left[ \log p^{\theta}(z_{\hat{t} : \hat{t} \geq t} \mid \mathbf{\beta}) \right] - \mathrm{KL}[{ q^{\phi}(\beta \mid m_t, z_t) } \mid\mid{ p^{\theta}(\beta) }]
\end{array}
\end{equation}
\end{prop}

Where $J(\rho)$ is the objective function for the acting policy-gradient agent~\cite{lowe2019pitfalls}.

\begin{equation}
    \begin{array}{cc}
         &  J(\rho) = J_{pol}(\theta^\mathbf{a}) +  \lambda_{ent} J_{ent}(\theta^\mathbf{a}) +
\lambda_v J_{v}(\theta^\mathbf{v})
    \end{array}
\end{equation}\label{loss:cont}

Where $J_{pol}(\theta^\mathbf{a}) = \mathbb{E}_{\pi_{\theta^\mathbf{a}}} [-\log \pi_{\theta^\mathbf{a}}(a|o) \cdot (r - V_{\theta^\mathbf{v}}(o))]$ is the REINFORCE update for the action policy, and $V(o)$ is the learned value function to reduce variance and $\rho=\{\theta^\mathbf{a},\theta^\mathbf{v} \}$. $J_{v}$ is the value function mean squared error loss and $J_{ent}$ is an entropy term to encourage exploration. $\lambda_{ent}$ and $\lambda_{v}$ are scalar hyper-parameters.

\subsubsection{Speaker}\label{sec:cc}

Positive signalling is defined as a positive correlation between the speaker's observation and the corresponding message it sends, i.e., the speaker should produce similar messages when in similar situations. Various methods exist to measure positive signalling, such as speaker consistency, context independence, and instantaneous coordination \cite{lowe2019pitfalls}. However, there is an absence of methods that \emph{promote} positive signalling. We propose Concept-Clustering (CC), in which the speaker is encouraged to maintain consistency between messages and visual information. 

\begin{algorithm}
  \caption{Concept Clustering}\label{alg:cc}
  \begin{algorithmic}[1]
    \Procedure{CC}{$o$}\Comment{Algorithm for calculating Concept Clustering}
      \State $\overline{p}_{soft} \gets 0$ \Comment{pass any decoder function $f^{o|m}$.}
      \State $MSE \gets 0$ \Comment{pass the senders message network excl. softmax layer $\pi_m$}
      \For{$o \text{ in } BatchObs$} 
        \State \ $p_{soft} = Softmax^{\tau}(\pi_m(o) + \epsilon),  \ \epsilon \sim \mathcal{G}umbel(0,1)$
    \State \ $m_t = OneHot(argmax(p_{soft})) - StopGradient(p_{soft}) + p_{soft}$ 
    \State \ $\hat{o} = f^{o|m}(m_t)$ \Comment{Decode message into an observation}
    \State \ $\overline{p}_{soft}\ \pluseq \frac{p_{soft}}{\text{BatchSize}}$
    \State \  $MSE\ \pluseq \sum \frac{(\hat{o} - o)^2}{\text{BatchSize}*\text{\#Pixels}}$\Comment{ Where the max error can be 1.0 } 
      \EndFor
      \State \textbf{return} $CC =\sum \overline{p}_{soft} \log(\overline{p}_{soft}) + MSE$ 
    \EndProcedure\Comment{ The entropy term aids $\mathcal{G}\mathcal{S}$ sample exploration}
  \end{algorithmic}
\end{algorithm}

This is directly inspired from the Observation model $P(o_{t} \mid  z_t)$ from Table~\ref{table:definitions}, re-interpreted for a speaker agent as $P(o_{t} \mid  m_t)$, where $m_t \sim P(m_t | o_{t})$ is the speakers own message.

$\pi^{s}_m$ denotes a function which outputs the speaker agents message policy logits. $f^{O|m}$ is any decoder architecture which reconstructs the observation conditioned on a message, $BatchObs^s$ is a batch of sender observations. The softmax temperature parameter is $\tau$. Intuitively the CC loss encourages diversity across the set of possible messages, clustering the observations using a reconstruction loss.







%% file: 05_experiments.tex
\section{Experiments}

We evaluate our models on a set of navigation tasks in a 2-dimensional gridworld (Fig~\ref{fig:game-simple}).  The speaker receives an observation of the entire map, while the listener can only observe one pixel in any direction. Each map contains a forking path, one of which contains a flag. Which path contains the flag is randomly assigned, and the listening agent is given only enough time to fully explore one path, thus requiring an effective communication protocol to achieve greater than 50$\%$ success.

We repeated each experiment a minimum of three times, allowing each experiment to run for 200,000 episodes. We plot the 95\% confidence interval for each experiment.

\begin{figure}[h]
\begin{subfigure}{.32\linewidth}
  \centering
  \includegraphics[width=1.0\linewidth]{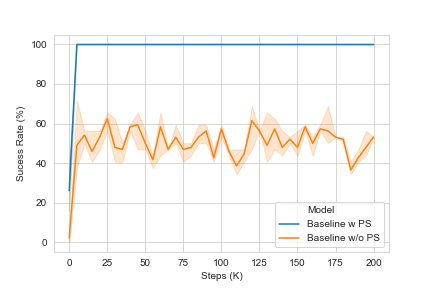}
\caption{Simple Game: Listener does not move, but guesses flag location.}
  \label{fig:sfig11}
\end{subfigure} \hfill 
\begin{subfigure}{.32\linewidth}
  \centering
  \vspace{-1em}
  \includegraphics[width=1.0\linewidth]{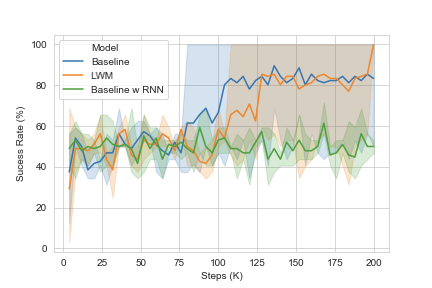}
\caption{Small action search space.}
\label{fig:sfig12}
\end{subfigure} \hfill 
\begin{subfigure}{.32\linewidth}
  \centering
  \vspace{-1em}
  \includegraphics[width=1.0\linewidth]{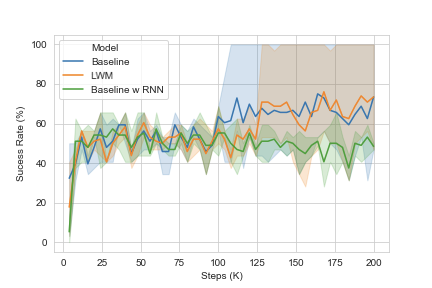}
\caption{Larger action search space.}
\label{fig:sfig13}
\end{subfigure}
\caption{(a) evaluates the use of CC loss to promote positive signalling on a classification game, (b) compares the performance of our LWM framework to existing methods, with and without our positive signalling loss, (c) compares the same models on a game with a longer horizon (i.e the flag placed further away from the agent). }
\end{figure}

\subsection{Evaluating Positive Signalling}

Fig~\ref{fig:sfig11} shows that even in this simple game, without the positive signalling loss to help promote the speaker's message consistency, the baseline models are unable to produce a useful communication protocol which solves the task.  
Similarly, Fig~\ref{fig:sfig12} shows the LWM and the baseline model have comparable performance. The increase of success above the non-communicative agent (\textasciitilde 50\% success), supports the claim that the agents learn effective communication, 
while we find the recurrent baseline model (LSTM) is unable to surpass the performance of a non-communicative agent. We believe this is due to an increase in parameters introduced to the learning problem. Fig~\ref{fig:sfig13} Shows a dramatic drop in performance when the game complexity is increased.

\begin{figure}[h!]
  \centering
  \includegraphics[width=.68\linewidth,trim=4 4 4 4,clip]{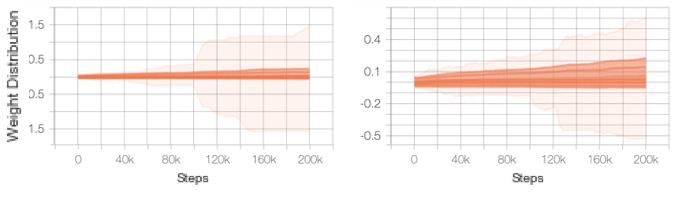}
  \caption{Value distribution of linear weight's during training, specifically, the listener agents communication channel (first layer units).}
  \label{fig:sfig1}
\end{figure}
\begin{figure}[h!]
  \centering
  \includegraphics[width=.68\linewidth,trim=4 4 4 4,clip]{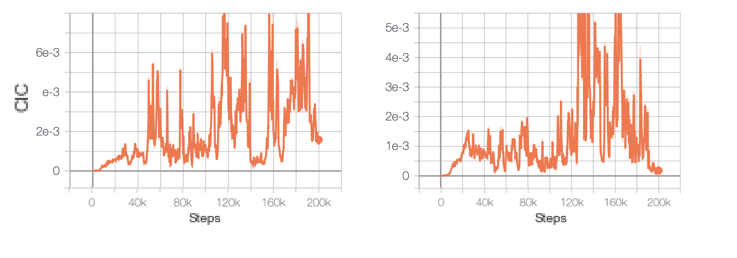}
  \caption{CIC measure during training of the LWM.}
  \label{fig:sfig2}
\end{figure}

\subsection{Evaluating Positive Listening}

As shown previously~\cite{lowe2019pitfalls}, positive signalling does not imply positive listening. We provide two additional sources of proof for positive listening in our LWM agent. Fig~\ref{fig:sfig1} supports the claim that our listener satisfies the positive listening criterion, as we see a widening of the input communication weights when the reward increases above the non-communicative agent (\~{}$50\%$ success), suggesting it has become more sensitive to the communication channel.  Fig~\ref{fig:sfig2} provides further support for positive listening, as we directly plot the CIC metric. CIC returning a value 1.0 implies the listener agents policy is extremely sensitive to communication. 

\subsection{Visualizing grounding}

\begin{figure}[h!]
\begin{subfigure}{.20\textwidth}
  \centering
  \includegraphics[width=1.0\linewidth]{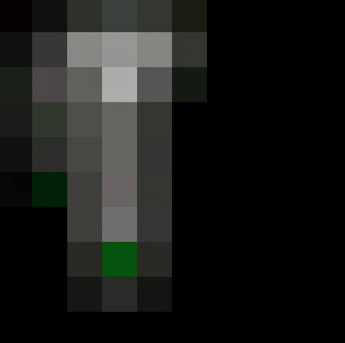}
  \caption{end of right corridor}
  \label{fig:sfig1}
\end{subfigure} \hfill 
\begin{subfigure}{.20\textwidth}
  \centering
  \includegraphics[width=1.0\linewidth]{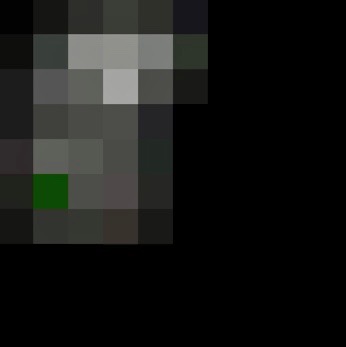}
  \caption{end of left corridor \ \ \ \ }
  \label{fig:sfig2}
\end{subfigure}\hfill 
\begin{subfigure}{.20\textwidth}
  \centering
  \includegraphics[width=1.0\linewidth]{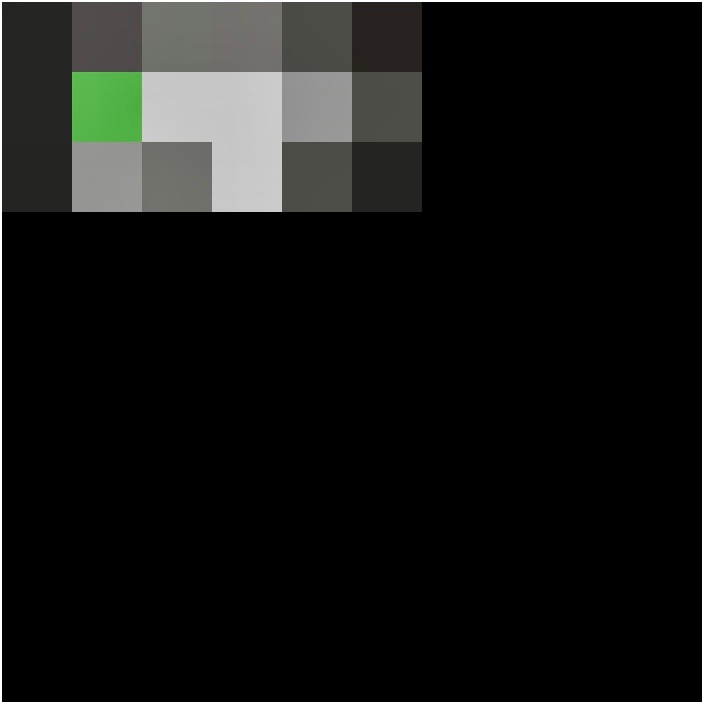}
  \caption{top of left corridor\ \ \ \ }
  \label{fig:sfig1}
\end{subfigure}\hfill 
\begin{subfigure}{.20\textwidth}
  \centering
  \includegraphics[width=1.0\linewidth]{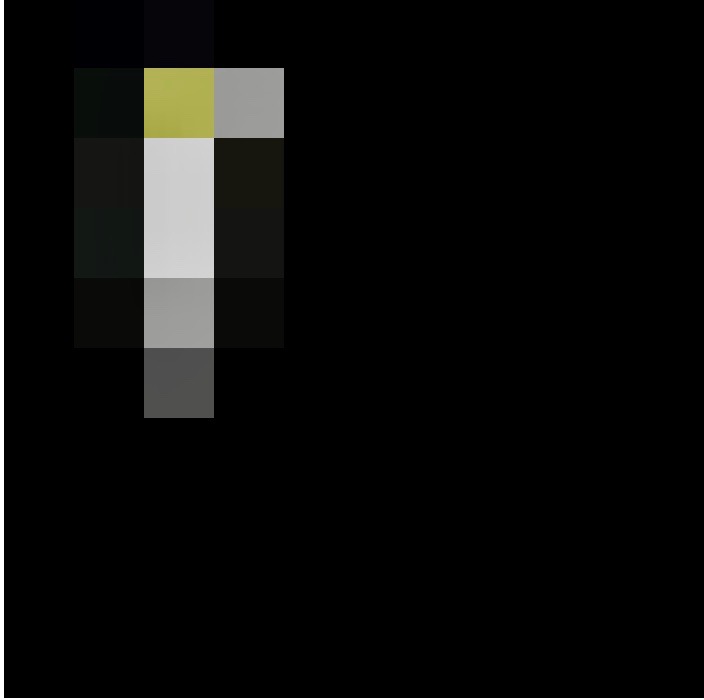}
  \caption{top of right corridor}
  \label{fig:sfig2}
\end{subfigure}
\caption{Visualisations of the LWM belief $\beta$, conditioned on a discrete message token and latent observation code $\beta \sim q^{\phi}(\beta \mid z_{t},m_{t})$, using the decoder $q^{\delta}$ directly on the belief code $\beta$. We enhanced saturation and exposure for clarity. An analysis of learned visualizations throughout various stages of optimizations is provided in the appendix.}
\label{fig:bviz}
\end{figure}

Fig~\ref{fig:bviz} sheds light on the type of message-conditional representations which have been incorporated into the belief state, a visual depiction of the learned message semantics. We see a visual depiction of the partial trajectory associated with the success in the task scenario, as conveyed by the message. In these figures the world is stationary, and only the location of the flag differs.  Therefore, the learned semantics of each message is quite coarse-grained, but it is expected that additional permutations in world state will hone message semantics into the specific object of interest.

We believe that as the community begins to pursue EC with more complex tasks, such methods for grounding messages in a persistent memory will become increasingly more valuable.

%% file: 06_related.tex
\section{Related Work}

There are two largely studied classes of learning algorithms in the field of emergent communication, with the training scheme either \emph{decentralised training} \cite{jaques2019social, foerster2016communication, singh2018learning, mordatch2018emergence} or \emph{centralised training} \cite{foerster2016communication}, both with \emph{decentralised acting}. In our work we focus on utilising a \emph{decentralised training} regime. 
A recent focus has been on the language developed, such as investigating emergent compositional language \cite{bogin2018emergence, choi2018compositional}, emergent referential language \cite{lazaridou2018referentialGame}, large scale multi-agent communicative systems\cite{Das2018targetMAC}.

A typical method for promoting emergence of communication is through encoding biases via auxiliary losses, such as using Casual Influence of Communication \cite{jaques2019social, lowe2019pitfalls}. Instead of adding an auxiliary loss to the policy, we continuously ground our language in observations. It has also been shown that grounding the messaging is a way of combating language drift \cite{lee2018countering}. 
\par 
There is a wealth of work in agents using a persistent memory for tasks \cite{gregor2018temporal, weston2014memory, graves2014neural, graves2016hybrid}. The work of \cite{gregor2018temporal} is most relevant, as their memory unit also contains a belief of future and past states, that is learned through ``jumpy prediction``, which breaks the single-step transition modelling limitation of typical recurrent models and allow modelling distant, temporally separated points. Other forms of multi-agent belief exist, such as public belief\cite{foerster2018bayesian}.

%% file: 07_conclusion.tex
\section{Discussion}

In the same way humans collect knowledge from a myriad of sources, from language and observation alike, only to compose facts and ultimately act upon them at a later time, we argue for the need to develop more persistent information in EC. We introduce the Language World Model as a way to accomplish this in a manner that is compatible with the strict constraints of EC work, and that is grounded in a domain which is amenable to available supervision. This allows us to develop effective auxiliary loss functions for guiding the emergence of effective communication.

This is not to discount the importance of current trends in EC, including EC in which communication more closely resembles the commander/follower paradigm we cited as a motivation for developing our method. Indeed, many simple EC games, including the ones presented here, may be solved equally well in such a paradigm, Here the distinction between ``\emph{go left}'' and ``\emph{the flag is down the left corridor}'' is not currently an important one. Yet to develop complex multi-agent strategy in games like Capture-the-Flag or Counter-Strike, it is likely that a combination of both directive and object-property communication will be necessary. We present this work as a first step towards these goals. 

In future work, we aim to extend this method to more complex, multi-object scenarios, where we relax some of the assumptions necessary for success in the current work. For instance, extending this work to scenarios where the reward function is not as deeply connected with the target of message grounding, where objects are dynamic, and where the visual grounding is not a projection of the observation alone, but may require additional steps of reasoning.  


\section{Acknowledgements}

We would like to thank Yuta Tsuboi, Sosuke Kobayashi, Prabhat Nagarajan, and the anonymous reviewers for helpful discussion and feedback on the draft of this work.

%% file: appendix.tex
\clearpage 
\section*{Appendix}



\begin{figure}[h]
\begin{subfigure}{.18\textwidth}
  \centering
  \includegraphics[width=1.0\linewidth]{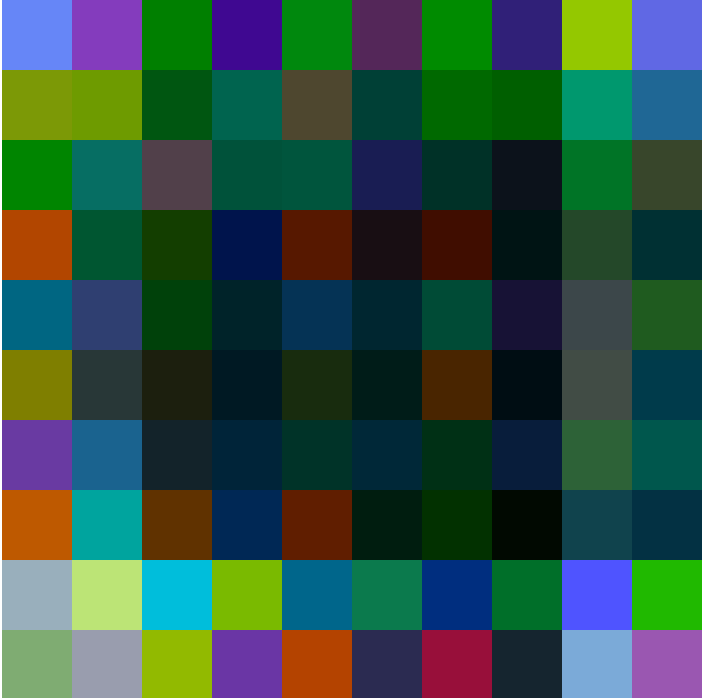}
\caption{5,000 steps} \end{subfigure} \hfill 
\begin{subfigure}{.18\textwidth}
  \centering
  \includegraphics[width=1.0\linewidth]{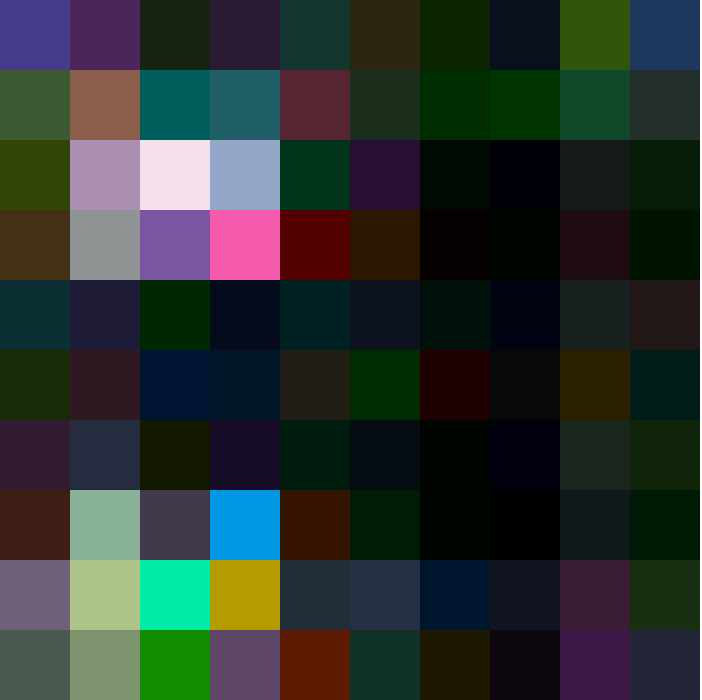}
\caption{10,000 steps}  \end{subfigure}\hfill 
\begin{subfigure}{.18\textwidth}
  \centering
  \includegraphics[width=1.0\linewidth]{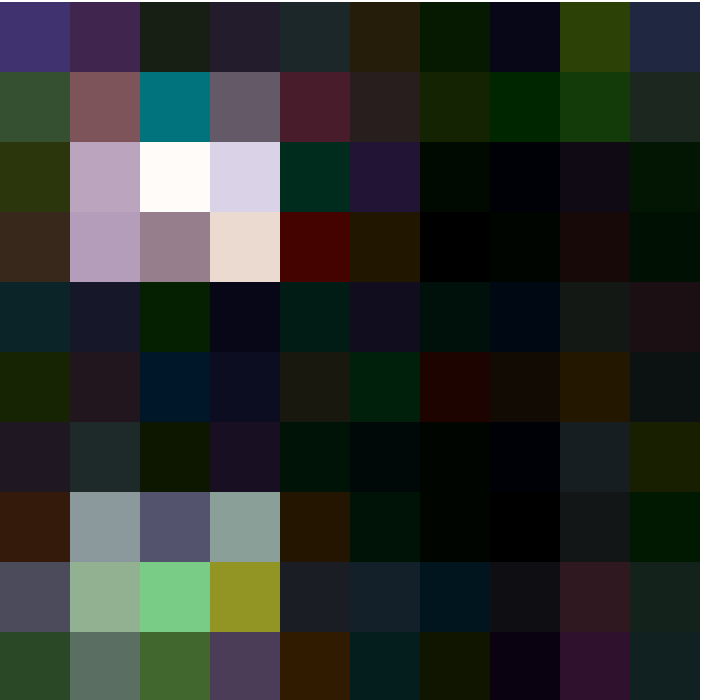}
\caption{15,000 steps}  \end{subfigure}\hfill 
\begin{subfigure}{.18\textwidth}
  \centering
  \includegraphics[width=1.0\linewidth]{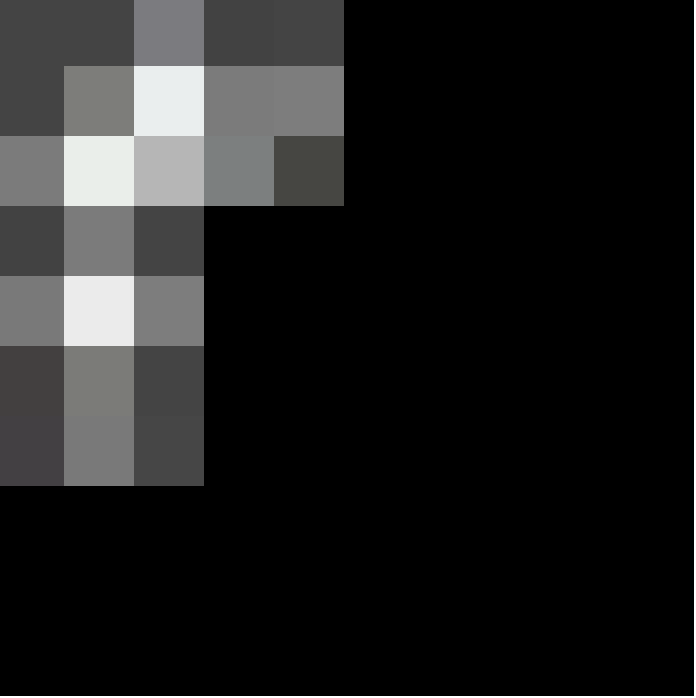}
\caption{20,000 steps}  \end{subfigure}
\begin{subfigure}{.18\textwidth}
  \centering
  \includegraphics[width=1.0\linewidth]{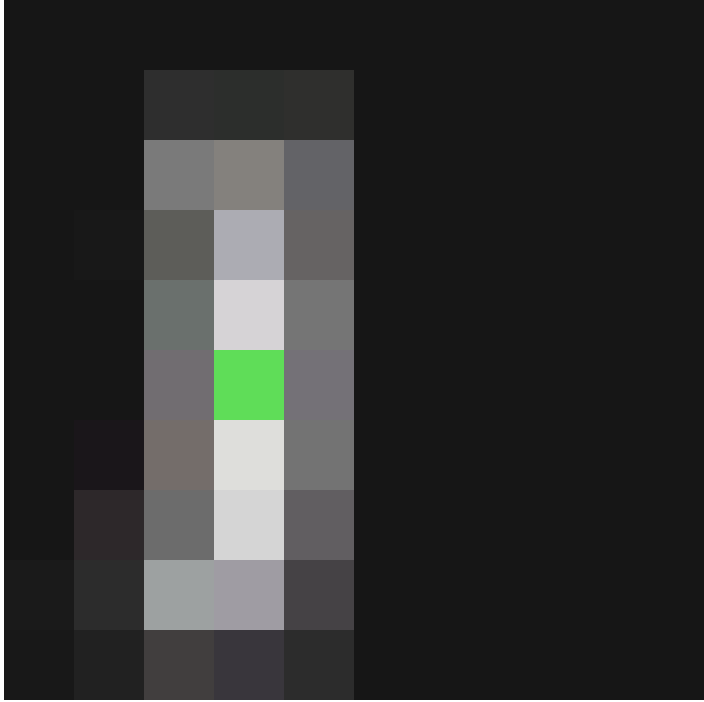}
\caption{25,000 steps}  \end{subfigure}
\begin{subfigure}{.18\textwidth}
  \centering
  \includegraphics[width=1.0\linewidth]{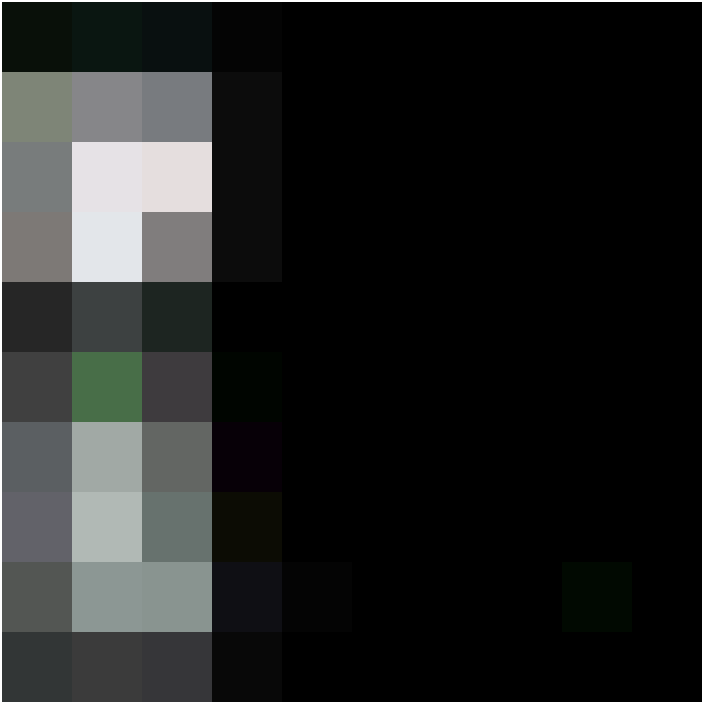}
\caption{30,000 steps}  \end{subfigure} \hfill 
\begin{subfigure}{.18\textwidth}
  \centering
  \includegraphics[width=1.0\linewidth]{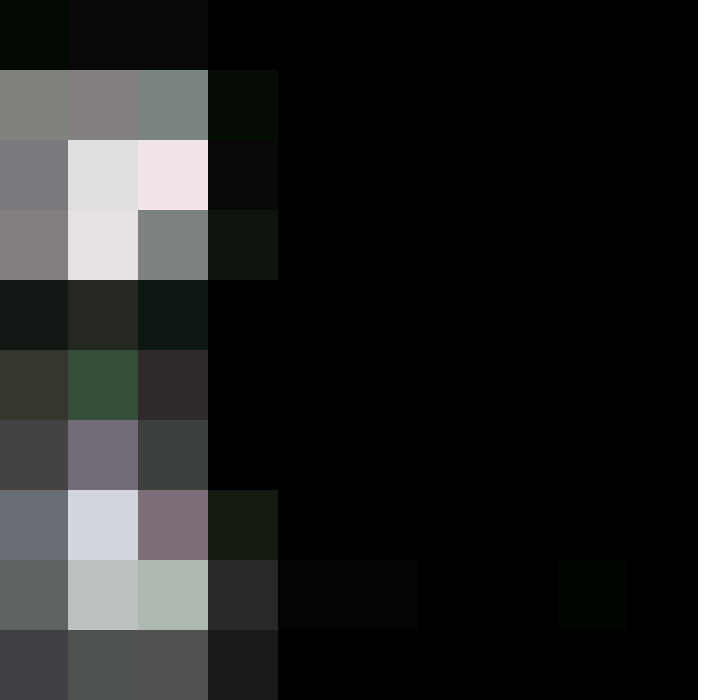}
\caption{35,000 steps}  \end{subfigure}\hfill 
\begin{subfigure}{.18\textwidth}
  \centering
  \includegraphics[width=1.0\linewidth]{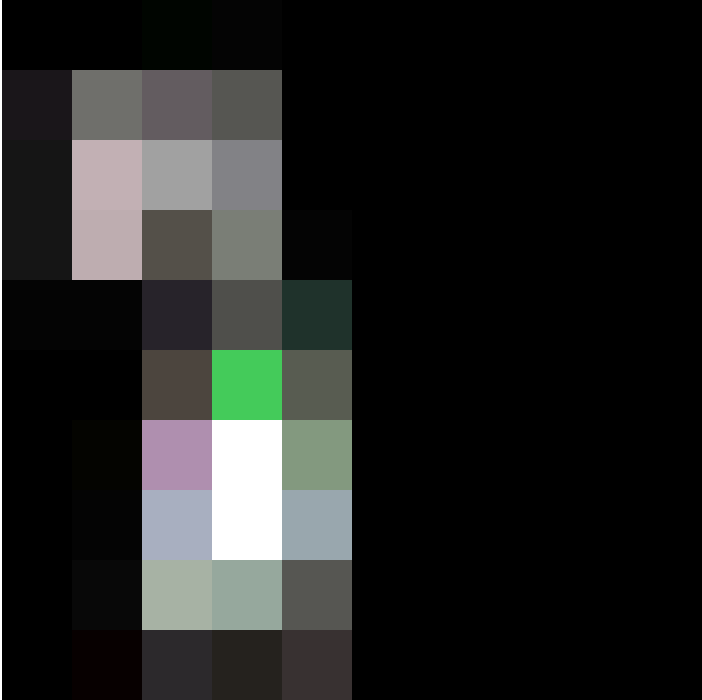}
\caption{40,000 steps}  \end{subfigure}\hfill 
\begin{subfigure}{.18\textwidth}
  \centering
  \includegraphics[width=1.0\linewidth]{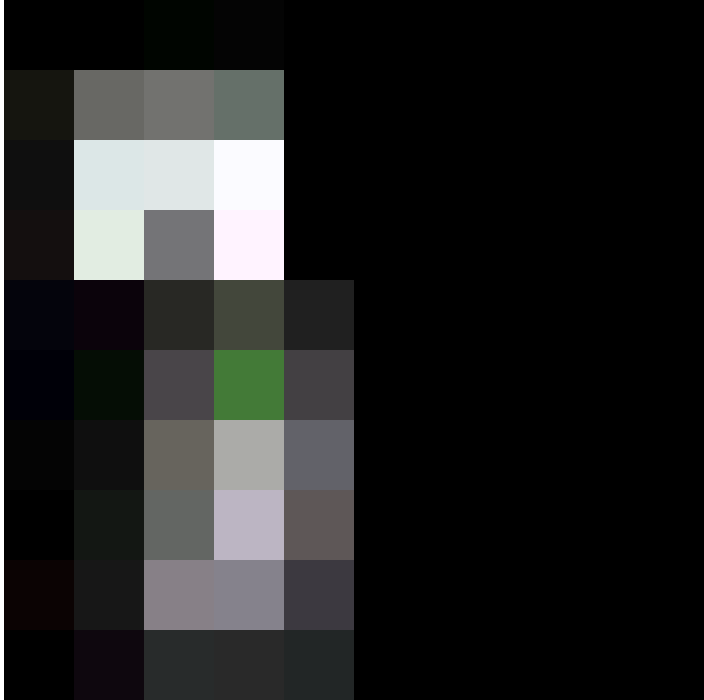}
\caption{45,000 steps}  \end{subfigure}
\begin{subfigure}{.18\textwidth}
  \centering
  \includegraphics[width=1.0\linewidth]{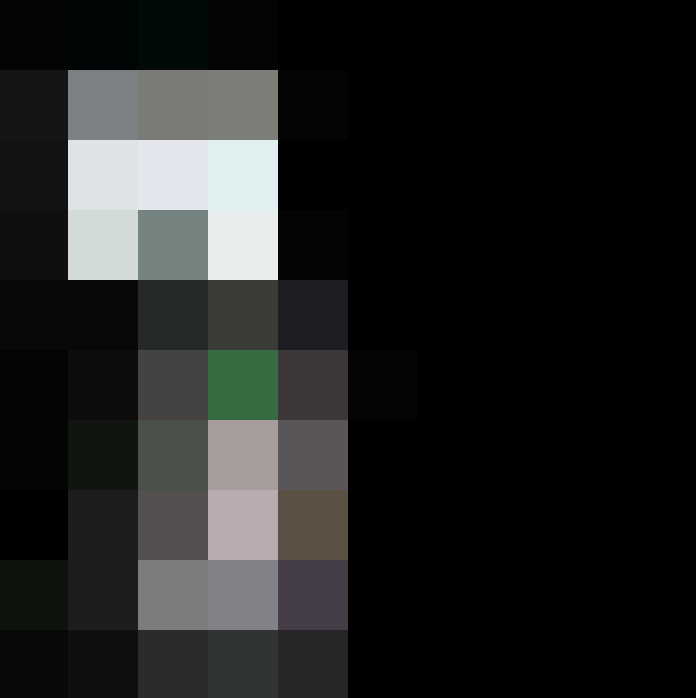}
\caption{50,000 steps}  \end{subfigure}
\begin{subfigure}{.18\textwidth}
  \centering
  \includegraphics[width=1.0\linewidth]{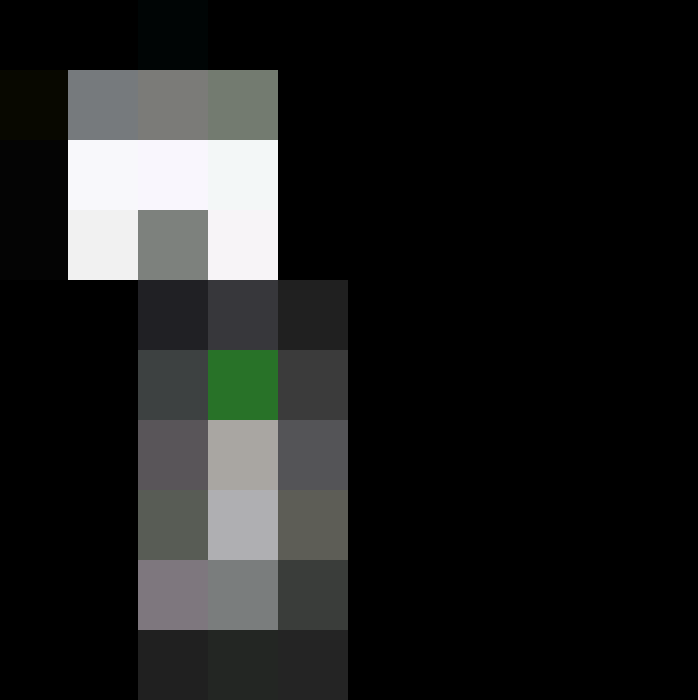}
\caption{55,000 steps} \end{subfigure} \hfill 
\begin{subfigure}{.18\textwidth}
  \centering
  \includegraphics[width=1.0\linewidth]{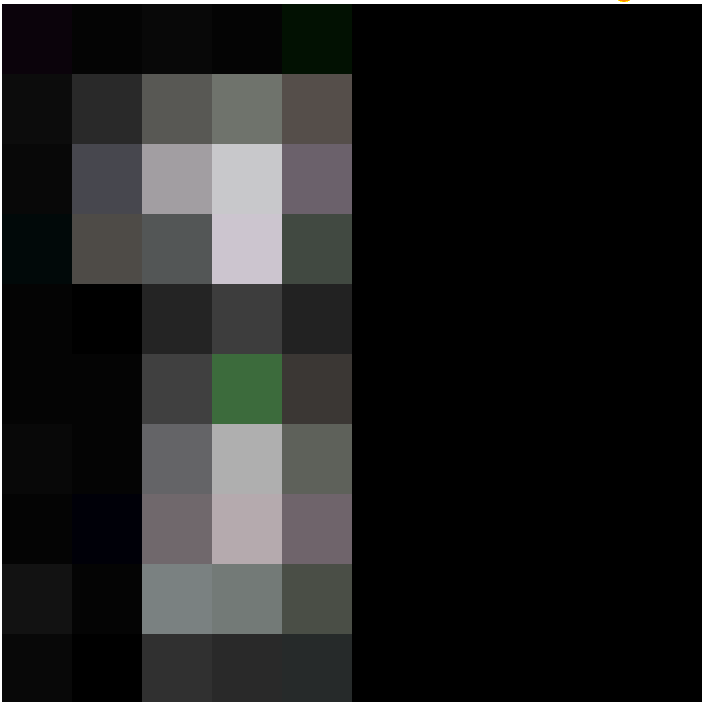}
\caption{60,000 steps}  \end{subfigure}\hfill 
\begin{subfigure}{.18\textwidth}
  \centering
  \includegraphics[width=1.0\linewidth]{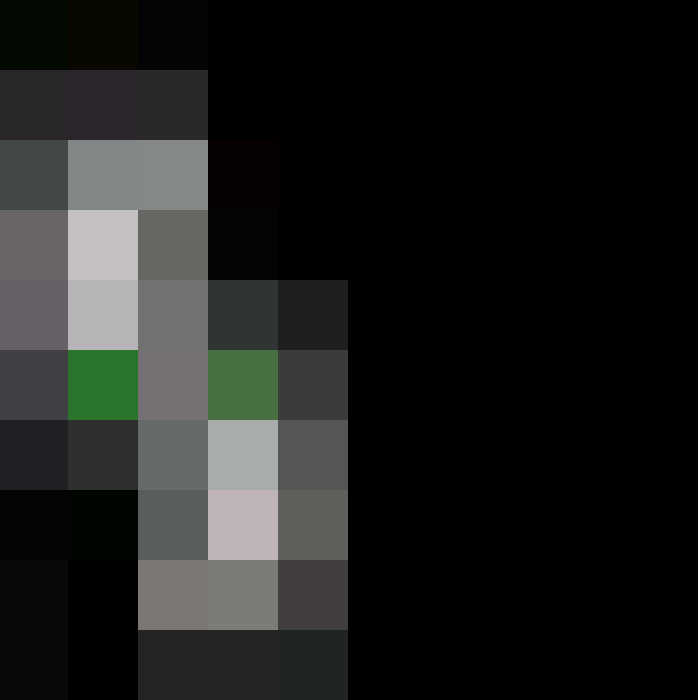}
\caption{65,000 steps}  \end{subfigure}\hfill 
\begin{subfigure}{.18\textwidth}
  \centering
  \includegraphics[width=1.0\linewidth]{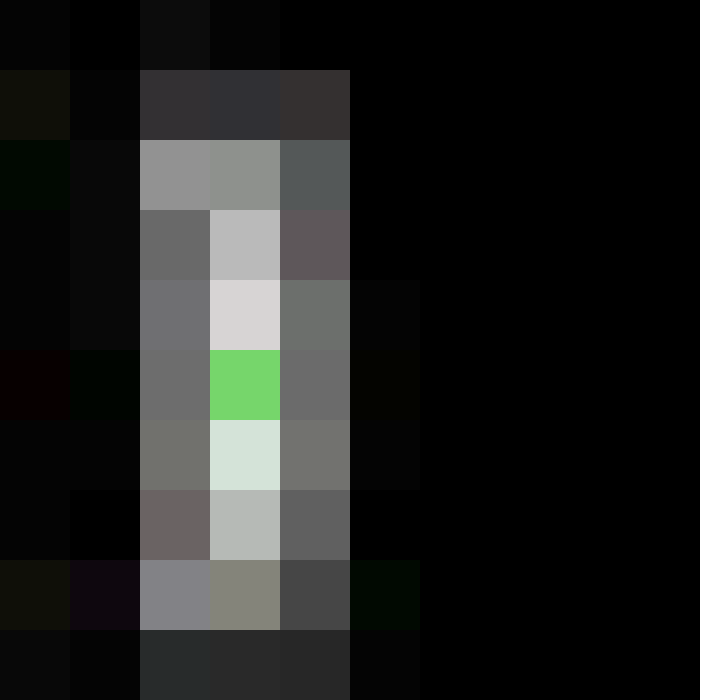}
\caption{70,000 steps}  \end{subfigure}
\begin{subfigure}{.18\textwidth}
  \centering
  \includegraphics[width=1.0\linewidth]{images/train/11142507.png}
\caption{75,000 steps}  \end{subfigure}
\caption{Visualisations of LWM's belief $\beta$, conditioned on a fixed discrete message token and latent observation code $\beta \sim q^{\phi}(\beta \mid z_{t},m_{t}=2)$, using the decoder $q^{\delta}$ directly on the belief code $\beta$. We enhanced saturation and exposure for clarity. We track one agent belief state over training, whilst fixing the message we visualise the belief over, in order to see how the message tokens representation/ belief changes over time. (a)-(c) we can see the agent is just learning the visual representation of the game, where (d) we can see the agent hasn't quite learned to link the message with the flags location, which is first seen in (e), whereby the agent first successfully learns the message token is connected visually with the flag in the rightmost corridoor.   }
\label{fig:trainbviz}
\end{figure}